\pdfoutput=1   
\documentclass[11pt]{article}

\PassOptionsToPackage{table}{xcolor}

\usepackage[preprint]{acl}

\usepackage{times}
\usepackage{latexsym}

\usepackage[T1]{fontenc}
\usepackage[utf8]{inputenc}

\usepackage{microtype}

\usepackage{inconsolata}

\usepackage{graphicx}
\usepackage{booktabs}
\usepackage{multirow}
\usepackage{enumitem}
\usepackage{amsmath}
\usepackage{amssymb}
\usepackage{listings}
\usepackage{xcolor}
\usepackage[most]{tcolorbox}
\newtcolorbox{evbox}{colback=gray!5, colframe=gray!60, boxrule=0.3pt, arc=1pt,
  left=3pt, right=3pt, top=1pt, bottom=1pt, before skip=3pt, after skip=3pt}
\newcommand{\evtext}[1]{\parbox{\linewidth}{\fontsize{6.6pt}{7.6pt}\selectfont
  \setlength{\parskip}{0pt}\setlength{\lineskiplimit}{-\maxdimen}#1}}
\newcounter{finding}
\newtcolorbox{findingbox}{colback=blue!3, colframe=blue!35, boxrule=0.5pt, arc=1.5pt,
  left=5pt, right=5pt, top=3pt, bottom=3pt, before skip=7pt, after skip=5pt}
\newcommand{\finding}[1]{%
  \refstepcounter{finding}%
  \begin{findingbox}\textbf{Finding \thefinding.} #1\end{findingbox}}
\usepackage{float}   
\usepackage{tabularx}
\newcolumntype{Y}{>{\centering\arraybackslash}X}  
\definecolor{rowshade}{gray}{0.88}
\definecolor{harmred}{HTML}{C0392B}   
\definecolor{safeblue}{HTML}{1F6FC4}  
\lstdefinestyle{prompt}{basicstyle=\footnotesize\ttfamily,breaklines=true,
  breakindent=0pt,columns=fullflexible,frame=single,framesep=3pt,
  xleftmargin=2pt,xrightmargin=2pt,aboveskip=4pt,belowskip=4pt,
  literate={—}{{-{}-}}2 {“}{{``}}1 {”}{{''}}1 {’}{{'}}1}

\title{Should We Type or Talk to LLM Agents?\\
A Comprehensive Study of Voice and Keyboard Input Perturbations}

\author{
  Zizhao Hu\textsuperscript{1} \quad
  Nathan Elijah Segura\textsuperscript{2} \quad
  Mohammad Rostami\textsuperscript{3} \quad
  Jesse Thomason\textsuperscript{1} \\[2pt]
  \textsuperscript{1}University of Southern California \quad
  \textsuperscript{2}Santa Monica College \\
  \textsuperscript{3}Information Sciences Institute, University of Southern California \\[2pt]
  \texttt{\{zizhaoh, jessetho\}@usc.edu}
}

\begin{document}
\maketitle

\begin{abstract}
Human input reaches language models by typing or speaking, and each channel leaves a
distinct signature: orthographic noise for keyboards; for voice, disfluency from
conventional transcription and restructuring from AI-backed dictation tools. How do they impact an LLM's performance? In this paper we present \textbf{HIVE} (Human
Input-Variation Engine), a suite of voice transcription perturbations and
QWERTY keyboard perturbations. We use HIVE to
evaluate how robust models are to these perturbations. We present seven findings. (i) Voice transcription perturbations
lower accuracy across every instruction-tuned model we test, and it is the structure of the
transcription rather than its fillers that carries the cost. (ii) QWERTY keyboard
perturbations cost less, and a model absorbs a lot of them before accuracy falls away.
(iii) Both trace back to one cause, how many of the question's tokens survive the
perturbation: destroying a token is what hurts, while adding new ones alongside it costs
little. (iv) The gap between the two channels appears only where the answer must be
constructed or deduced; on multiple choice there is none. (v) The harm does not solely come from test-set
contamination. (vi) It cannot be trained away with lightweight adaptation. (vii) A
thinking budget recovers the keyboard channel almost entirely but leaves the spoken registers
untouched, and compressed speech is worse with it.
\end{abstract}

\section{Introduction}

\label{sec:intro}

Nearly all text that reaches a deployed language model is produced by a human
who is either typing or speaking, and neither channel delivers it clean. A QWERTY keyboard
introduces orthographic noise: neighbor-key substitutions, transpositions, dropped or
doubled characters, casing and punctuation slips. Speech reaches the model through a
transcription layer that leaves a different signature: preserved disfluencies (``um'',
``like'', false starts), homophone substitutions, word-boundary errors, and, increasingly,
a wholesale rewrite of what was said. These perturbations are systematic and benign
(the by-product of ordinary human input, not adversarial attacks), yet they
are exactly the inputs a model must tolerate in production.

Speaking is now a first-class path into a language model rather than a niche one.
Coding agents such as Claude Code and Codex accept dictated instructions. Phone
assistants such as Google Assistant and Siri route a spoken request to an LLM backend.
Dictation front-ends such as Typeless add a further LLM-powered layer that cleans and
reformats the user's spoken request before it is sent. In each case the model receives a
transcript, and in each case the speaker might not proofread the string that was actually
sent. Whatever the transcription pipeline leaves behind, or rewrites, is what the model
has to answer.

Robustness to typographical noise has been studied at length
\citep{belinkov2018synthetic,ebrahimi2018hotflip,pruthi2019combating,gan2024reasoning}, and so has robustness to verbal disfluency, through disfluent-question
corpora \citep{gupta2021disflqa} and, more recently, benchmarks that inject fillers,
hesitations, false starts and self-corrections into spoken queries
\citep{liu2025vocalbenchdf,lin2026fullduplexv3}. Two gaps remain. First, some of these works are
largely benchmarking. They establish that noisy input lowers scores, but leave
how each modality does so vague, so they yield little guidance a person could act on
when deciding how to address an AI agent. Second, the transcription layer between the two
literatures is untouched. Voice interfaces now hand the model an LLM-rewritten version of
what was said rather than a verbatim one, and no study we know of measures what that
rewriting costs. We instead treat the QWERTY keyboard and the AI-based
transcription model as the two most common sources of input a deployed system receives, and
focus on the perturbations native to each: the motor errors a keyboard actually produces,
and the disfluency and rewriting a transcription layer actually introduces.

Our contributions are fourfold.

\paragraph{Contribution 1: we build a perturbation suite for real human input errors.} \textbf{HIVE},
seventeen operators and two controls built from LLM style transfer and deterministic rules
(\S\ref{sec:hive}, Table~\ref{tab:suite}).

\paragraph{Contribution 2: we measure how much input perturbation costs LLM task accuracy, per channel.}
Findings~\ref{sec:f1}, \ref{sec:f2} and \ref{sec:f4}.

\paragraph{Contribution 3: we explain where the damage from input perturbation comes from.}
Finding~\ref{sec:fcause}.

\paragraph{Contribution 4: we show what does and does not mitigate the harm from input perturbation.}
Findings~\ref{sec:f5}, \ref{sec:ftrain} and \ref{sec:fthink}.

\section{Related Work}

\label{sec:related}

\paragraph{LLM input perturbation robustness.} A large literature perturbs input text and
measures the resulting drop, organised largely by the granularity of the edit.
Character-level work injects misspellings and typographical errors
\citep{gao2018blackbox,li2019textbugger,pruthi2019combating,gan2024reasoning}, building on
earlier findings that adversarial and natural misspellings degrade neural models
\citep{belinkov2018synthetic,ebrahimi2018hotflip} and on behavioral test suites such as
CheckList \citep{ribeiro2020checklist}, which includes typo invariance tests. Word-level work
substitutes synonyms or shuffles word order \citep{garg2020bae,jin2020bert,moradi2021evaluating}. Sentence-level work paraphrases the input or
inserts irrelevant context \citep{shi2023distracted,lanham2023measuring}, and
even minor punctuation noise moves LLM accuracy \citep{abedin2025arithmattack}. These edits are largely theoretical, chosen to probe a
model rather than to reproduce anything a user does. Our textual perturbations instead bind
to the QWERTY keyboard and the mistakes ordinary typing produces on it, so the damage we
measure is the damage a deployed system actually receives.

\begin{figure*}[t!]
\centering
\includegraphics[width=\textwidth]{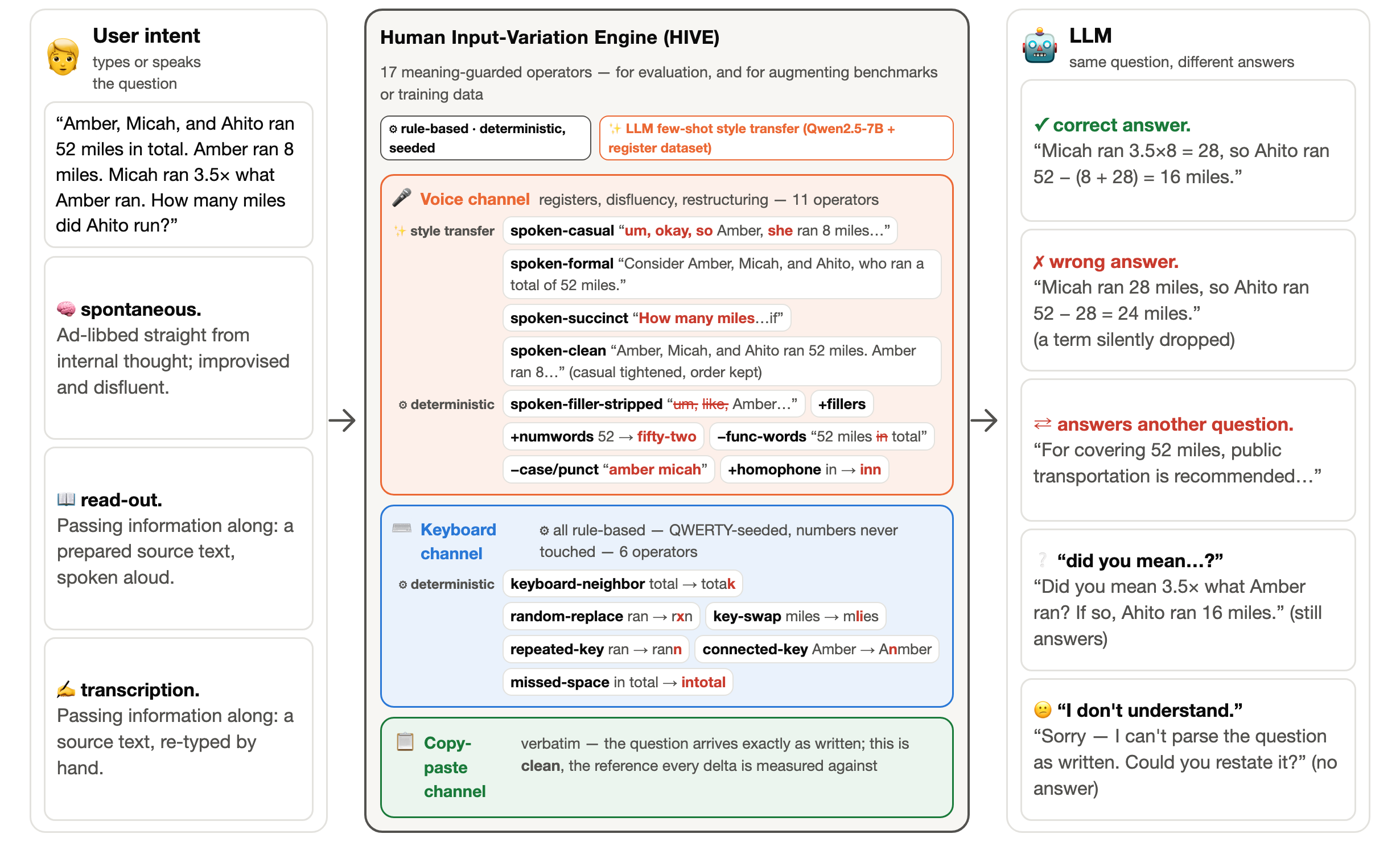}
\caption{\textbf{The HIVE taxonomy at a glance.} A user either ad-libs a question or passes
existing information along, and it reaches the model through voice, a QWERTY keyboard, or
copy-paste, which delivers the question verbatim and is the clean reference. Colour marks the channel, not how an operator is built:
the voice operators are a mix of few-shot LLM style transfer and deterministic rules, and the
keyboard operators are all deterministic. Every example shown is a real item from the
released data. The right panel shows how a model replies under
degraded input. Table~\ref{tab:methods} gives the full specification.}
\label{fig:hive_taxonomy}
\end{figure*}

\paragraph{Voice and transcription robustness.} Downstream tasks have been evaluated over
ASR transcripts for some time, through spoken question answering \citep{lee2018spokensquad} and voice-assistant understanding corpora \citep{bastianelli2020slurp,fitzgerald2022massive}, with disfluency characterised separately in conversational speech
\citep{godfrey1992switchboard} and disfluent-question corpora \citep{gupta2021disflqa}. Recent work measures how transcription errors propagate into a downstream task: they
derail code understanding by dropping underscores and substituting phonetically similar
words for code terms \citep{havare2026lostintranscription}, degrade spoken question answering
comparably across LLMs of very different ability \citep{jung2026korean}, and can
be partly trained against with synthetic ASR noise \citep{binici2024medsage}. Throughout, the
transcription layer is treated as noise to be corrected rather than as an input register in
its own right, so these studies do not isolate different spoken registers, and none compares
speaking against the typing channel the same user could have chosen. Nor do they study the
case where the input is rewritten by an AI-based transcription model, so what an LLM
receives from a voice interface is not what the user said but a rewrite of it.

\section{HIVE}
\label{sec:setup}\label{sec:hive}

\paragraph{The HIVE engine.} Every input change we test comes from one instrument, the
Human Input-Variation Engine (Figure~\ref{fig:hive_taxonomy}). It models a user who
either ad-libs a question straight from internal thought or passes along text they
already have, and delivers it through one of three channels: \textbf{voice}
(11 operators, \S\ref{sec:voiceperturb}), \textbf{keyboard} (6 operators,
\S\ref{sec:kbd}), and \textbf{copy-paste}, the question verbatim, which is clean and the
reference every delta is measured against. Two further operators are experimental
controls rather than input channels: reordering context and question, and permuting
answer options. Each operator is implemented either as a seeded rule-based transform or
as few-shot LLM style transfer (Qwen2.5-7B prompted with a register dataset);
Table~\ref{tab:methods} specifies all of them with a worked example.

\subsection{Voice transcription perturbations}
\label{sec:voiceperturb}
Eleven operators model what reaches a model when the user speaks. Two \textbf{registers}
rewrite the stem as a person would say it, conversational or prepared-talk. Three
\textbf{rewrites} model the LLM cleanup layer that dictation tools apply, either compressing
the utterance or tightening its wording while preserving clause order. One
\textbf{filler-strip} deletes hesitations and changes nothing else. Five \textbf{deterministic
factors} isolate single surface properties a transcript carries: injected fillers, numbers as
words, dropped function words, lost casing and punctuation, and homophone substitution. The
registers and rewrites are produced by LLM style transfer with a local Qwen2.5-7B-Instruct
model, so no paired written/spoken data is needed and the operators apply to any benchmark
with a text question. The split matters: a register is what a person said,
a rewrite is what a dictation tool sent on their behalf, and only the second is under a
developer's control.

\subsection{QWERTY keyboard perturbations}
\label{sec:kbd}
Six operators model what reaches a model when the user types, each the textual residue of a
distinct motor error rather than an arbitrary edit: a neighbouring key struck instead of the
intended one, two adjacent letters transposed, a letter duplicated, an adjacent letter struck
alongside the intended one, a clipped space merging two words, and, as the control with no
human behind it, a letter replaced at random. Edits are applied over the
Damerau--Levenshtein basis at a fixed per-word distance on a fixed share of eligible words,
seeded by item id, with numeric tokens protected so the gold answer stays valid.
Appendix~\ref{sec:opspec} defines each operator and Table~\ref{tab:methods} gives its knobs
and a worked example.

\begin{table*}[!t]
\centering
\footnotesize
\renewcommand{\arraystretch}{1.02}
\setlength{\aboverulesep}{0.2ex}\setlength{\belowrulesep}{0.2ex}
\setlength{\tabcolsep}{2pt}
\begin{tabularx}{\textwidth}{@{}l *{6}{Y} !{\vrule width 0.5pt} YY@{}}
\toprule
Operator & GSM8K & GSM-Sym & GSM1k & MMLU-Pro & TruthfulQA & HumanEval & GM & GM$_{\!j}$ \\
\midrule
\emph{Baseline} \quad clean, accuracy \% & $71.8$\,{\tiny$\pm$24.2} & $67.9$\,{\tiny$\pm$25.9} & $65.1$\,{\tiny$\pm$22.5} & $32.6$\,{\tiny$\pm$10.5} & $58.6$\,{\tiny$\pm$6.9} & $69.9$\,{\tiny$\pm$17.3} & $59.0$ & --- \\
\midrule
\multicolumn{8}{@{}l}{\emph{Controls \quad {\footnotesize (rows below: change vs.\ clean, points)}}} \\[1pt]
context/question swap & $-2.0$\,{\tiny$\pm$3.3} & $-1.5$\,{\tiny$\pm$2.3} & $-2.5$\,{\tiny$\pm$3.4} & n/a & n/a & n/a & $-3.0$ & $-23.3$ \\
option-permutation & n/a & n/a & n/a & $+0.6$\,{\tiny$\pm$2.8} & $-0.4$\,{\tiny$\pm$3.5} & n/a & $+0.6$ & $+0.6$ \\
\midrule
\multicolumn{8}{@{}l}{\emph{Voice transcription perturbations}} \\[1pt]
spoken-casual & $-5.7$\,{\tiny$\pm$3.3} & $-5.1$\,{\tiny$\pm$3.0} & $-5.1$\,{\tiny$\pm$3.9} & $-1.6$\,{\tiny$\pm$3.3} & $-4.1$\,{\tiny$\pm$3.1} & $-6.2$\,{\tiny$\pm$7.7} & $-7.4$ & $-7.0$ \\
spoken-formal & $-2.6$\,{\tiny$\pm$3.1} & $-5.4$\,{\tiny$\pm$3.1} & $-2.4$\,{\tiny$\pm$3.2} & $-1.7$\,{\tiny$\pm$2.1} & $-1.2$\,{\tiny$\pm$3.1} & $-1.5$\,{\tiny$\pm$3.9} & $-4.1$ & $-3.7$ \\
spoken-succinct & \textcolor{harmred}{$-26.3$\,{\tiny$\pm$8.2}} & \textcolor{harmred}{$-24.7$\,{\tiny$\pm$10.3}} & \textcolor{harmred}{$-22.6$\,{\tiny$\pm$9.1}} & $-1.3$\,{\tiny$\pm$4.1} & $-2.6$\,{\tiny$\pm$3.9} & \textcolor{harmred}{$-14.4$\,{\tiny$\pm$7.9}} & \textcolor{harmred}{$-24.1$} & \textcolor{harmred}{$-14.9$} \\
spoken-clean & $-6.6$\,{\tiny$\pm$3.0} & $-6.1$\,{\tiny$\pm$3.2} & $-4.4$\,{\tiny$\pm$3.9} & $-1.2$\,{\tiny$\pm$3.2} & $-2.5$\,{\tiny$\pm$2.9} & $-6.3$\,{\tiny$\pm$6.2} & $-7.0$ & $-5.5$ \\
spoken-clean (Llama) & $-6.4$\,{\tiny$\pm$3.5} & $-5.4$\,{\tiny$\pm$3.5} & $-4.0$\,{\tiny$\pm$3.5} & $-1.8$\,{\tiny$\pm$3.6} & $-2.2$\,{\tiny$\pm$2.7} & $-11.0$\,{\tiny$\pm$7.8} & $-8.1$ & $-6.4$ \\
spoken-filler-stripped & $-6.2$\,{\tiny$\pm$3.1} & $-5.8$\,{\tiny$\pm$3.6} & $-4.1$\,{\tiny$\pm$3.6} & $-1.3$\,{\tiny$\pm$3.3} & \textcolor{harmred}{$-5.0$\,{\tiny$\pm$3.9}} & $-6.7$\,{\tiny$\pm$6.1} & $-7.6$ & $-6.7$ \\
clean+fillers & $-3.4$\,{\tiny$\pm$3.1} & $-2.0$\,{\tiny$\pm$3.1} & $-1.6$\,{\tiny$\pm$3.2} & \textcolor{harmred}{$-1.9$\,{\tiny$\pm$2.5}} & $-3.6$\,{\tiny$\pm$3.4} & $-3.0$\,{\tiny$\pm$7.8} & $-4.4$ & $-4.3$ \\
clean+numwords & \textcolor{safeblue}{$-0.5$\,{\tiny$\pm$2.7}} & $-1.0$\,{\tiny$\pm$3.3} & $-0.3$\,{\tiny$\pm$3.5} & $-0.7$\,{\tiny$\pm$1.8} & \textcolor{safeblue}{$+0.3$\,{\tiny$\pm$1.4}} & $-5.1$\,{\tiny$\pm$5.0} & $-2.0$ & $-0.4$ \\
clean$-$func-words & $-0.7$\,{\tiny$\pm$1.8} & $-0.4$\,{\tiny$\pm$2.6} & \textcolor{safeblue}{$+0.3$\,{\tiny$\pm$2.4}} & $+0.1$\,{\tiny$\pm$2.0} & $+0.3$\,{\tiny$\pm$2.2} & \textcolor{safeblue}{$-0.5$\,{\tiny$\pm$4.0}} & \textcolor{safeblue}{$-0.2$} & \textcolor{safeblue}{$-0.2$} \\
clean$-$case/punct & $-2.2$\,{\tiny$\pm$3.1} & $-1.3$\,{\tiny$\pm$2.5} & $-1.9$\,{\tiny$\pm$3.3} & $-0.1$\,{\tiny$\pm$2.9} & $-0.8$\,{\tiny$\pm$2.9} & $-4.5$\,{\tiny$\pm$3.9} & $-2.7$ & $-2.7$ \\
clean+homophone & $-1.0$\,{\tiny$\pm$2.3} & \textcolor{safeblue}{$-0.3$\,{\tiny$\pm$2.0}} & $-0.0$\,{\tiny$\pm$1.9} & \textcolor{safeblue}{$+0.5$\,{\tiny$\pm$1.2}} & $-0.4$\,{\tiny$\pm$1.5} & $-0.8$\,{\tiny$\pm$1.2} & $-0.4$ & $-0.4$ \\
\midrule
\multicolumn{8}{@{}l}{\emph{QWERTY keyboard perturbations}} \\[1pt]
random-replace & \textcolor{harmred}{$-5.1$\,{\tiny$\pm$4.2}} & \textcolor{harmred}{$-5.0$\,{\tiny$\pm$3.3}} & $-2.2$\,{\tiny$\pm$2.2} & \textcolor{harmred}{$-1.8$\,{\tiny$\pm$2.5}} & \textcolor{harmred}{$-4.8$\,{\tiny$\pm$3.1}} & \textcolor{harmred}{$-2.5$\,{\tiny$\pm$4.9}} & \textcolor{harmred}{$-5.9$} & \textcolor{harmred}{$-5.3$} \\
keyboard-neighbor & $-4.2$\,{\tiny$\pm$3.3} & $-4.1$\,{\tiny$\pm$2.7} & \textcolor{harmred}{$-2.7$\,{\tiny$\pm$2.8}} & $-1.4$\,{\tiny$\pm$2.6} & $-4.1$\,{\tiny$\pm$2.8} & $-1.5$\,{\tiny$\pm$4.0} & $-5.0$ & $-4.6$ \\
key-swap & $-1.9$\,{\tiny$\pm$3.2} & $-1.5$\,{\tiny$\pm$2.9} & $-1.1$\,{\tiny$\pm$2.4} & $-1.3$\,{\tiny$\pm$2.4} & $-2.8$\,{\tiny$\pm$3.1} & $-1.1$\,{\tiny$\pm$3.6} & $-2.8$ & $-2.3$ \\
repeated-key & \textcolor{safeblue}{$-0.6$\,{\tiny$\pm$2.5}} & \textcolor{safeblue}{$+0.1$\,{\tiny$\pm$2.4}} & $-0.0$\,{\tiny$\pm$3.6} & $-0.7$\,{\tiny$\pm$2.1} & $-1.3$\,{\tiny$\pm$2.2} & $-2.0$\,{\tiny$\pm$3.5} & $-1.4$ & $-1.3$ \\
connected-key & $-1.3$\,{\tiny$\pm$2.9} & $-0.2$\,{\tiny$\pm$3.1} & \textcolor{safeblue}{$+0.4$\,{\tiny$\pm$3.0}} & $-1.4$\,{\tiny$\pm$2.2} & $-3.0$\,{\tiny$\pm$2.6} & $-1.3$\,{\tiny$\pm$4.8} & $-2.1$ & $-1.9$ \\
missed-space & $-1.0$\,{\tiny$\pm$3.0} & $-0.2$\,{\tiny$\pm$1.8} & $+0.3$\,{\tiny$\pm$2.5} & \textcolor{safeblue}{$-0.7$\,{\tiny$\pm$2.6}} & \textcolor{safeblue}{$-1.3$\,{\tiny$\pm$2.3}} & \textcolor{safeblue}{$+0.3$\,{\tiny$\pm$2.0}} & \textcolor{safeblue}{$-0.9$} & \textcolor{safeblue}{$-0.8$} \\
\bottomrule
\end{tabularx}
\caption{\textbf{Full perturbation suite.} Accuracy change vs.\ clean, in percentage
points, mean\,{\tiny$\pm$}\,s.d.\ over the 25 (model, seed) cells; the first row is the
clean baseline in absolute accuracy (\%). \textbf{GM} is the geometric mean of the
per-benchmark accuracy ratios, so it is a relative change; \textbf{GM$_{\!j}$} recomputes it on the items an LLM judge scores as still asking the same question. Within each modality block and column the
\textcolor{harmred}{most damaging} operator is red and the \textcolor{safeblue}{least
damaging} blue. \emph{n/a}: the operator does not
apply to that benchmark.}
\label{tab:suite}
\end{table*}

\section{Experiments}
\label{sec:expsettings}

\paragraph{Models, benchmarks, seeds.} We evaluate five instruction-tuned models
(Llama-3.1-8B, Qwen2.5-7B, Mistral-7B-v0.3, Qwen3-8B, phi-4) at five seeds on six
benchmarks: GSM8K \citep{cobbe2021gsm8k}, GSM-Symbolic \citep{mirzadeh2024gsmsymbolic} and
GSM1k \citep{zhang2024gsm1k} (free-form arithmetic), HumanEval \citep{chen2021humaneval}
(Python synthesis, pass@1 by unit test), and MMLU-Pro STEM \citep{wang2024mmlupro} and
TruthfulQA MC1 \citep{lin2022truthfulqa} (multiple choice). We draw $n{=}200$ items per benchmark ($n{=}164$, the full set, for
HumanEval), fixed across conditions by shared item ids. Decoding is greedy, so a
matched pair isolates the perturbation rather than sampling noise. Crossing this with
the seventeen operators and two controls gives 550k scored generations. Every item appears once per
condition and is scored against its own clean counterpart in the same (model, seed) cell,
so every change we report is a within-item comparison.

\paragraph{Test-set contamination controls.} GSM-Symbolic and GSM1k are rebuilds of the
GSM8K task carrying the same operators, so any effect that is really memorization recovery
should shrink on them (Finding~\ref{sec:f5}). Operators, prompts and statistics are
specified in \S\ref{sec:method}.

\section{Main Results}
\label{sec:results_main}

\begin{figure*}[t!]
\centering
\includegraphics[width=\textwidth]{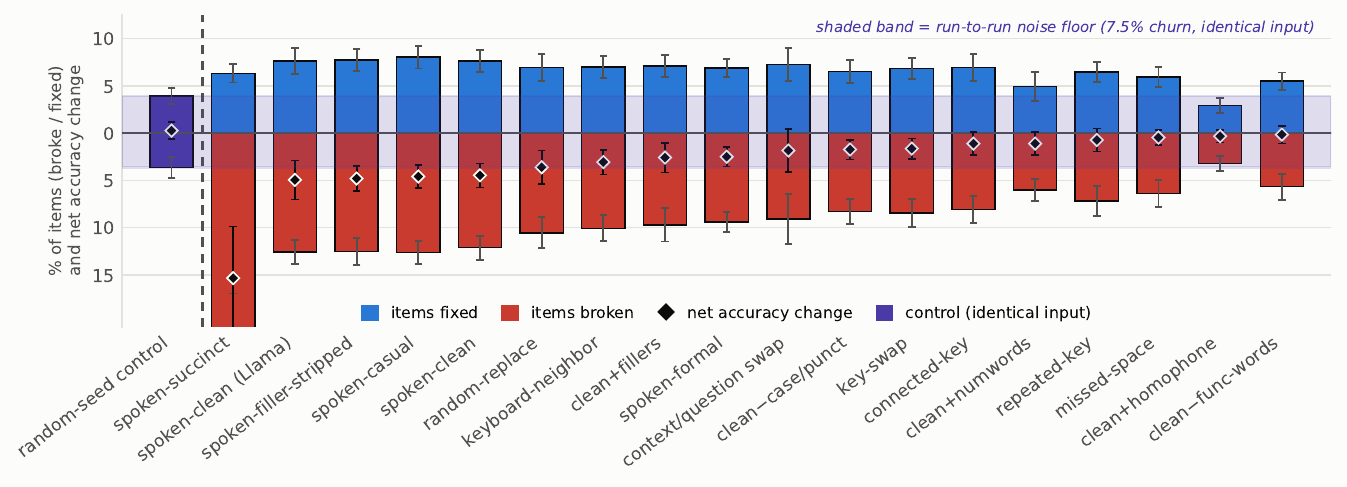}
\caption{\textbf{Every input change we test, and what each one is made of.} For each
operator, the blue bar is the share of items it fixed, the red bar the share it broke, and
the black diamond their difference, exactly the net accuracy change an aggregate table
reports. Means over the 25 (model, seed) cells; whiskers are $\pm1$ s.d.
\textbf{The leftmost bar is a zero-perturbation control}: the same clean item scored twice,
which flips $7.5\%$ of answers purely from run-to-run nondeterminism, an artifact of batched left-padded decoding and non-deterministic reduction order rather than sampling. That churn is the
shaded band, the floor every operator must clear.}
\label{fig:suite_overview}
\end{figure*}

\begin{figure*}[!t]
\centering
\includegraphics[width=\textwidth]{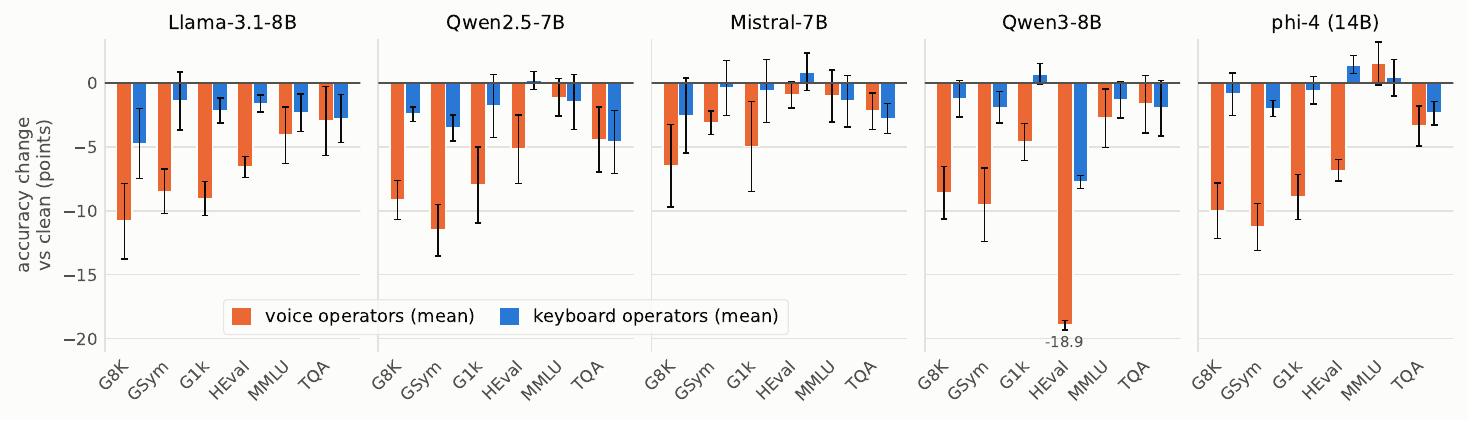}
\caption{\textbf{Per-model granularity.} Accuracy change vs.\ clean in percentage points,
one panel per
model; error bars are the s.d.\ across five seeds. Voice sits below keyboard in all 20
generation cells; the two interleave on MMLU-Pro and TruthfulQA.}
\label{fig:model_granularity}
\end{figure*}

The full suite is reported in Table~\ref{tab:suite}, broken out model by model in
Figure~\ref{fig:model_granularity}, and decomposed into the items each operator breaks
and fixes in Figure~\ref{fig:suite_overview}. The first row is the clean baseline and every row below it a
change against it on the same items, so a cell reads as percentage points lost. Every operator has a negative or negligible mean effect and none
helps. The spread inside each block is wide, $-24.1$ to
$-0.2$ on the voice side and $-5.9$ to $-0.9$ on the keyboard side, so the findings below
read the blocks operator by operator. GM$_{\!j}$ changes little except for the compressing
rewrite, the one operator that also risks changing what was asked.

\section{Findings}
\label{sec:type_or_talk}

\finding{Voice transcription perturbations cost accuracy across instruction-tuned models, and their structure rather than their fillers carries the cost}
\label{sec:f1}

The six spoken rows of Table~\ref{tab:suite} cost more than any other block:
compressing the request into textbook form $-24.1$ (Table~\ref{tab:suite}), tightening its wording while keeping
clause order $-8.1$ and $-7.0$ for two different rewriters, stripping the fillers out of the
transcription $-7.6$, a conversational transcription $-7.4$, and a prepared-talk
transcription $-4.1$, a mean of $-9.7$ (geometric mean of the per-benchmark accuracy ratios;
$6.2$ raw points). The registers order
themselves by how far they depart from written form, and the prepared register is about
half as damaging as the conversational one, so what matters is not that the input was spoken
but how it was spoken. Under GM$_{\!j}$ the block mean holds at $-7.4$ and
compression stays the most damaging operator by a wide margin. Homophone substitution is the exception that costs nothing at all
($-0.4$), though with at most three swaps per stem and number-words excluded this may reflect
the operator's gentleness as much as any robustness to homophones.

\paragraph{Fillers are sufficient but not necessary for the spoken penalty.} Two
deterministic controls bracket the disfluency hypothesis, and they are the two rows of
Table~\ref{tab:suite} that differ from the conversational transcription in exactly one
respect. Adding fillers to an otherwise clean written question ($-4.4$) leaves accuracy
$2.6{\pm}1.5$ percentage points lower,
so fillers alone are enough to hurt. But deleting them from the spoken transcript
($-7.6$) recovers nothing at all: $-4.80{\pm}1.35$ against
the untouched transcription's $-4.59{\pm}1.24$, a paired difference of $-0.22{\pm}1.08$ over
25 (model, seed) cells ($t{=}{-}1.00$, n.s.), if anything slightly worse. If fillers carried the penalty, removing them
would return the accuracy; it does not.

\finding{QWERTY keyboard perturbations cost less, and models absorb a lot of them before breaking}
\label{sec:f2}

The keyboard block is milder than the spoken one throughout. It spans $-5.9$
(replacing a letter at random) to $-0.9$ (dropping a space) and averages $-3.0$ against the
spoken block's $-9.7$ (Table~\ref{tab:suite}). The blocks overlap at the boundary: the single worst
typing error, replacing a letter at random at $-5.9$, is more damaging than the mildest
transcription, the prepared-talk one at $-4.1$. Typing is not uniformly safer than
speaking, though the typical typed input is.

\begin{figure*}[t]
\centering
\includegraphics[width=\textwidth]{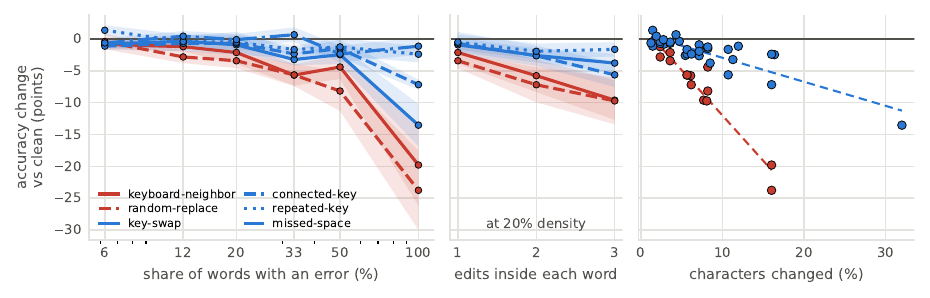}
\caption{\textbf{How much of a perturbation a model absorbs.} Accuracy change vs.\ clean in
percentage points, pooled over 2 models $\times$ 3 benchmarks; bands are $\pm1$ s.e.\ over
those cells. Red operators replace an
original letter; blue ones leave every letter present. \textbf{Left}: how many words carry an
error. \textbf{Middle}: how many edits land inside each word. \textbf{Right}: the share of
characters the edit changed, with a fit per family. The families lie on different lines, so
edit distance predicts damage within a family but not across them.}
\label{fig:dose_response}
\end{figure*}

\paragraph{QWERTY-shaped errors cost less than arbitrary ones.} The comparison that holds
edit type and density fixed and varies only whether the substituted character is a spatial
neighbor finds a real advantage. Hitting a neighbouring key leaves accuracy
$3.07{\pm}1.31$ percentage points lower against $3.61{\pm}1.79$ for an arbitrary letter, a paired difference of
$+0.55{\pm}1.15$ over the 25 cells ($t{=}2.37$, $p{<}0.05$). One explanation is that QWERTY separates
the letters that occur together most often, so a neighbour substitution is confined to pairs
the language rarely uses and less often forms a plausible alternative token. The margin is
small, so this is a conjecture. The four cheapest typing errors (transposing two adjacent
letters $-2.8$, hitting an extra key alongside the intended one $-2.1$, duplicating a letter
$-1.4$, dropping a space $-0.9$) are also the four that leave every
original letter in place, which Finding~\ref{sec:fcause} takes up.

\paragraph{Models absorb a surprising amount of typing error, then break suddenly.} We
re-ran each keyboard operator over a ladder of intensities, varying how many words carry an
error ($6\%$ to $100\%$) and how many edits land inside each ($1$ to $3$). The response is
not proportional (Figure~\ref{fig:dose_response}). Accuracy is flat out to roughly $12\%$ of
words and then falls away, reaching $-30.5$ and $-39.7$ on GSM8K once every word is hit.
Concentration matters more than spread: three edits inside $20\%$ of the words is about
twice as damaging as one edit spread over $50\%$, because a word edited three times is
unrecoverable while three singly-edited words each remain readable in context.

\begin{figure}[H]
\centering
\includegraphics[width=\columnwidth]{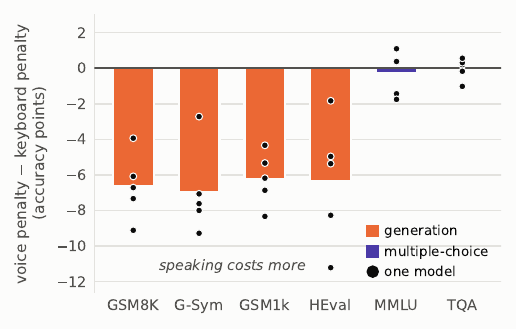}
\caption{\textbf{The modality gap is a construction-task effect.} Voice minus keyboard
penalty per benchmark (bars: mean over models; dots: the 5 models). All 20 generation cells
share the negative sign; on the multiple-choice sets the spread crosses zero.}
\label{fig:modality_gap}
\end{figure}

\finding{The modality gap is a reasoning-task effect: it appears only where the answer must be constructed or deduced}
\label{sec:f4}

\paragraph{Which modality you use only matters when the model has to build the answer.}
The two channels are not uniformly far apart. The gap between them tracks what the task
asks the model to do: on the three arithmetic sets and on code, where an answer has to be
reasoned out and written from scratch, speech is $6$--$7$ percentage points lower than
typing, with
the same sign in all $20$ model$\times$benchmark cells
(Figure~\ref{fig:modality_gap}). On the two multiple-choice sets, where the answer is
selected from options already supplied, the gap is $-0.3{\pm}1.3$ and $-0.0{\pm}0.6$,
standard deviations several times the means, so we report no difference there.
Absolute damage follows the same ordering: averaged over every operator, MMLU-Pro loses
$1.1{\pm}2.9$ points where GSM8K loses $4.3{\pm}6.8$ and HumanEval $5.3{\pm}13.6$.

\begin{figure}[!ht]
\centering
\includegraphics[width=\columnwidth]{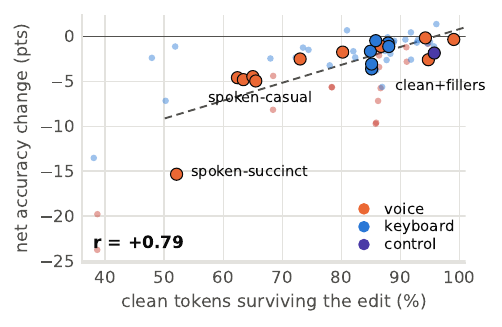}
\caption{\textbf{How much of the clean question's tokenization survives a perturbation.}
Large dots are operators, small dots the same keyboard operators at higher dose.}
\label{fig:token_survival}
\end{figure}

\finding{The shared cause of input perturbation harm is how many of the question's original tokens survive}
\label{sec:fcause}\label{sec:f3dose}%

\paragraph{Destroying the original tokens is what hurts. Adding new ones on top is close
to free.} An edit can do two separate things to the tokenized question. It can
\textbf{destroy} tokens that were there, replacing them with something the model has not
seen in that position, and it can \textbf{add} tokens that were not there before, making the
sequence longer without touching the originals. Tokenizing every clean/perturbed pair with
the evaluated model's own tokenizer lets us measure the two independently: what share of
the clean question's tokens still survive, and what share of the perturbed question is new.
Only destruction predicts the damage. Dropping the compressing rewrite, the operator
furthest from the rest, raises the correlation to $+0.82$ rather than lowering it, so the
relationship is not an artifact of one extreme. Survival correlates with the net effect at
$r{=}{+}0.79$ ($n{=}18$ operators, $p{<}0.001$), while addition correlates at only
$r{=}{-}0.28$ (Figure~\ref{fig:token_survival}). The extremes line up exactly. Compressing
the request destroys the most, leaving only $52.1\%$ of the original tokens intact, and is
the most damaging operator in the suite at $15.3$ percentage points lower. The two benign
operators leave $94$ to $99\%$ of the tokens intact and change accuracy by no measurable
amount (homophone substitution $99.0\%$ intact, $-0.33$; removing function words $94.2\%$,
$-0.15$). This refines the token-inflation probe of \S\ref{sec:mechanism},
which counted tokens without asking whether the original ones were still there.

\paragraph{Token survival is the dominant predictor, and it holds at the character level
too.}
Two operators damage while keeping their tokens. Injecting fillers keeps $94.7\%$ of the
clean tokens yet adds $46\%$ more and is still $2.6$ percentage points lower, dilution rather
than destruction; moving the question to the front keeps $95.7\%$ and adds only $4.0\%$, yet
is $1.8$ points lower, damage that is positional rather than lexical. The same rule holds on the raw string:
pooled over operators, Damerau--Levenshtein distance predicts poorly ($r{=}-0.63$) because
the two families lie on different lines, and at an identical $15.7\%$ of characters changed
replacing a letter costs $23.8$ points against $2.4$ for transposing or duplicating one.

\finding{The harm caused by input perturbation does not solely come from test-set contamination}
\label{sec:f5}

\paragraph{The same harm pattern holds on a decontaminated rebuild of GSM8K.}
GSM-Symbolic \citep{mirzadeh2024gsmsymbolic} regenerates GSM8K problems from symbolic templates, resampling the names and numbers so the
surface form is new while the reasoning is preserved. If spoken phrasing mainly blocked
retrieval of a solution the model had
already seen, the penalty should shrink once the strings are new. It does not. The voice
penalty is $-8.99$ on GSM8K against $-8.77$ on GSM-Symbolic, and the ordering of operators
within the block is unchanged (Table~\ref{tab:suite}, and per model in
Figure~\ref{fig:model_granularity}). GSM1k \citep{zhang2024gsm1k}, an independently authored set in the same style, gives
$-7.09$, so roughly $79\%$ of the effect persists on the furthest rebuild. The two controls
are not equally strong: template regeneration keeps the parent problem's structure, so clean
accuracy on GSM-Symbolic stays within $2$ points of GSM8K for three of the five models, while
the newly written GSM1k is $7.1$ to $8.5$ points lower for those same models. GSM-Symbolic is
therefore evidence that the harm is not tied to particular strings, and GSM1k the stronger
evidence that it is not tied to the problems either. We do not read the missing fifth as noise: some of it may genuinely be memorization recovery, which is why we claim contamination is not the main source of the harm rather than no source at all.

\finding{The harm caused by input perturbation cannot be trained away with lightweight adaptation}
\label{sec:ftrain}%

Can a model be taught to absorb the penalty rather than have its input cleaned? We ran self-distillation: for each item the model answers the clean question, we keep only
the items it gets right, and those answers become the target for the same model shown the
perturbed question. We distil a LoRA adapter on one seed's variants, evaluate on held-out seeds, and require that clean
accuracy not regress, since robustness bought by degrading the clean path is not robustness.
Three settings were swept (Table~\ref{tab:frsd}).

\paragraph{Adapting to the errors costs base performance.} The two regimes trade against
each other and neither wins. At high rank and learning rate the conversational-speech gap
does close, but only because every condition is dragged down to it. Clean falls from $60.9$
to $58.0$, which is the outcome the clean-accuracy constraint was meant to exclude. At low rank, one epoch and a
small learning rate, clean survives ($60.9 \to 60.3$) but conversational speech is untouched
($57.2 \to 56.5$). The sweep found no setting between the two. Any adapter strong enough to
move the perturbed condition moves the clean one further. Adding a term that penalises attention to filler tokens, or one that aligns the perturbed hidden state to the clean one, changes neither regime.

\begin{table}[H]
\centering
\footnotesize
\setlength{\aboverulesep}{0.3ex}\setlength{\belowrulesep}{0.3ex}
\setlength{\tabcolsep}{3pt}
\begin{tabularx}{\columnwidth}{@{}l YYY@{}}
\toprule
Llama-3.1-8B & Clean & Spoken- & Spoken- \\
(held-out)   &       & casual  & formal  \\
\midrule
Base (no adaptation)  & 60.9 & 57.2 & 60.2 \\
High rank             & 58.0 & 58.3 & 58.3 \\
Low rank              & 60.3 & 56.5 & 58.5 \\
Low rank + hidden-align & 60.2 & 56.5 & 58.2 \\
\bottomrule
\end{tabularx}
\caption{Nothing we tried recovers the conversational-speech penalty. Held-out
accuracy (\%).}
\label{tab:frsd}
\end{table}

\finding{Thinking recovers the keyboard channel almost entirely and leaves the spoken registers essentially untouched}
\label{sec:fthink}

Qwen3-8B is the one model we run both with its thinking block enabled and disabled, which
isolates the reasoning budget with the weights and perturbations held fixed. Both passes are
scored on the same $164$ GSM8K items, each referenced to its own pass's clean accuracy, since
thinking lifts clean from $78.7$ to $88.4$.

\begin{figure}[H]
\centering
\includegraphics[width=\columnwidth]{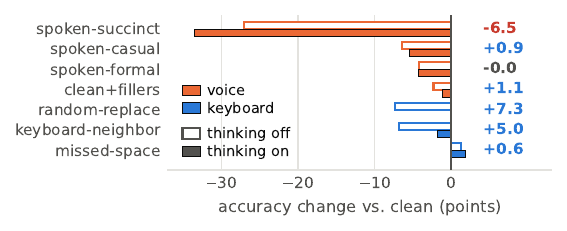}
\caption{\textbf{Thinking recovers the keyboard channel, not the spoken registers.}
Qwen3-8B on the same $164$ GSM8K items, thinking off (open bars) and on (solid), each
referenced to that pass's own clean.}
\label{fig:thinking_budget}
\end{figure}

\paragraph{Reasoning repairs a garbled question, not a simplified one.} Thinking absorbs the
keyboard operators almost entirely (Figure~\ref{fig:thinking_budget}): replacing a letter at
random goes from $7.3$ percentage points lower to no measurable change and hitting a
neighbouring key from $6.8$ to $1.8$, while injecting fillers goes from $2.3$ to $1.2$. The
spoken registers barely move, a conversational transcription from $6.4$ to $5.5$ and a
prepared-talk one from $4.3$ to $4.3$, and compression gets worse, $27.0$ points lower
without thinking and $33.5$ with it. The split follows Finding~\ref{sec:fcause}: the
operators reasoning recovers are the ones that leave the original tokens in place, and the
ones it cannot recover are the ones that destroy them. This is one model on one benchmark at
$164$ items, so we read the ordering, not the values.

\section{Conclusion}

In this work we present \textbf{HIVE}, a suite of seventeen operators and two controls that
reproduce the errors human input actually carries, and an analysis using it across five
instruction-tuned models and six benchmarks. We find that speaking is the expensive channel,
that what damage costs is how much of the question's original tokenization it destroys, and
that neither test-set contamination nor lightweight adaptation accounts for or repairs the
harm. Several takeaways follow. If you type, a reasoning model absorbs your errors. If you
dictate it does not, so the phrasing you speak is the phrasing the model works from. And if
you build dictation tools, do not reformat or restructure what the user said: strip
disfluency minimally and leave homophones alone. That last one matters most, because
dictation front-ends are becoming a default way to reach a model and their rewriting layer is
the largest harm we measure and the cheapest thing to change.

\section*{Limitations}

\paragraph{Scope.} Every benchmark is English, every model is an open 7--14B instruction-tuned
one, and every item is a single-turn question. Production voice interfaces route to far
stronger models, and our own Finding~\ref{sec:fthink} shows that a thinking budget absorbs
most of the suite, so the magnitudes here should not be read as estimates for a deployed
frontier assistant. Nothing in the suite exercises a multi-turn or tool-using agent.

\paragraph{One keyboard layout.} The typing arm models QWERTY only. Other layouts exist, and
error modes differ across devices: a phone thumb keyboard, a tablet, and a physical keyboard
do not fail in the same way. We chose QWERTY because it is the layout most people type on,
and because blind typing on it is where human supervision of the input is weakest. A user
who does not look at what was sent is the user whose raw errors reach the model intact, so
this is the setting that exposes an assistant to the most uncorrected input, and the
one layout worth measuring first is therefore also the one of greatest practical
consequence.

\paragraph{Synthetic speech, not audio.} Our spoken registers are produced by an LLM
verbalizer, not by a real speaker through a real recogniser, so they carry disfluency and
restructuring but not acoustic confusions or a true ASR error distribution. An acoustic branch, a
TTS$\to$ASR round-trip, is implemented in our harness but is not in the reported runs, and
not only for cost. Synthetic speech is too clean: a machine voice articulates perfectly, so a
recogniser transcribes it almost exactly and the resulting text differs barely at all from
what went in. A round-trip through synthetic audio therefore does not reproduce the errors a
human speaker produces, and would mostly measure the fidelity of the TTS system. Getting the
acoustic branch right needs recorded human speech, not a synthetic proxy. We accept this
trade deliberately. Because the operators
are text-to-text, they apply to any existing benchmark that has a written question, with no
audio to record, no speakers to recruit and no per-dataset pipeline to build, which is what
makes a suite of this size affordable and what lets the same operators be reused to run the
same analysis on other benchmarks. An audio-grounded study would measure a narrower slice of the
problem at far greater cost, and the two are complementary rather than substitutes.

\bibliography{custom}

\clearpage
\appendix

\section{Methodology, experiments and discussion}
\label{sec:appendix-method}

The main text is organised around the findings. This appendix carries the specification and analysis they rest on: how the operators, tasks, models and statistics are defined, how the runs were produced, and the discussion of what the results mean beyond the findings themselves.

\subsection{Methodology}
\label{sec:framework}\label{sec:method}

Let $x \in \Sigma^{*}$ be a clean benchmark item and $g(f(x)) \in \{0,1\}$ its
correctness, where $f$ is the model-plus-decoding map and $g$ the benchmark's
scorer. HIVE's core object is the perturbation operator $\tau_{m,c}$, which
maps $x$ to a perturbed item
$\tilde{x}$, indexed by a modality $m \in \{\mathrm{kbd}, \mathrm{voice}\}$ and a
condition $c$ (e.g.\ spoken-casual or spoken-formal for voice; a selection policy for keyboard).
Because each operator maps any clean text to its variation, the same suite applies
unchanged to any benchmark with a written question.
For item $i$ we form the \textbf{matched pair}
$\big(Y_{i,m,0},\, Y_{i,m,c}\big) = \big(g(f(x_i)),\, g(f(\tau_{m,c}(x_i)))\big)$,
the clean and perturbed correctness of the same item, and study the paired
difference; averaging over items gives the accuracy delta
$\Delta_{m,c} = \tfrac{1}{N}\sum_i \big(Y_{i,m,c} - Y_{i,m,0}\big)$.

Most operators are \textbf{intent-preserving}: a reader recovers the original
intent, whether from a non-word typo or a spoken disfluency, or, where the
surface token changes, from context (a real-word typo or a homophone). A small
\textbf{meaning-changing control} genuinely changes what is asked (a
multiple-choice option permutation that relabels the gold) and serves as an
over-robustness check. The
same object $(Y_{i,m,0}, Y_{i,m,c})$ is well defined for both modalities, which is
what lets one statistical apparatus (\S\ref{sec:stats}) serve both. The keyboard modality shares the identical definitions.

\subsection{Operator specification}
\label{sec:opspec}

Each operator of \S\ref{sec:voiceperturb} and \S\ref{sec:kbd} is defined below, numbered as
in Table~\ref{tab:methods}, which also gives its control knobs, its drift rate and a worked
example on a single running stem.

\paragraph{Voice transcription operators (Table~\ref{tab:methods}, rows 3--13).}
\textbf{Spoken-casual} (row 3) targets a conversational, podcast-like register at a high
filler rate ($\varphi \approx 5\%$), carrying fillers, false starts and self-corrections.
\textbf{Spoken-formal} (row 4) targets a prepared-talk register at $\varphi \approx 1.5\%$,
measured complete sentences with only occasional hesitations. Both are produced by LLM style
transfer: a local Qwen2.5-7B-Instruct model rewrites the written stem under a prompt carrying
the target register, a filler lexicon and two few-shot written$\to$spoken exemplars, listed
verbatim in Appendix~\ref{sec:prompts}. \textbf{Spoken-succinct} (row 5) compresses the
transcript into concise textbook form, which also reorders it. \textbf{Spoken-clean} (rows
6--7) tightens the wording while preserving the original clause order and question position,
isolating verbosity from reordering; it is run under two different rewriters so the effect is
not one model's habit. \textbf{Spoken-filler-stripped} (row 8) is a delete-only pass over the
spoken transcript that removes filler tokens and tidies the resulting punctuation, leaving
every content word, number and relationship exactly as spoken. We use a deterministic strip
rather than an LLM editor on purpose, since an editor would conflate the intervention with
its own paraphrase errors. Rows 9--13 are the deterministic factors, each isolating one
surface property of a transcript with no LLM involved: \textbf{clean+fillers} injects
hesitations into the otherwise untouched written question, \textbf{clean+numwords} writes
every quantity as words, \textbf{clean$-$func-words} drops articles and prepositions,
\textbf{clean$-$case/punct} lowercases and strips punctuation, and
\textbf{clean+homophone} substitutes acoustically confusable words.

\paragraph{QWERTY keyboard operators (Table~\ref{tab:methods}, rows 16--21).}
All six edit the stem over the Damerau--Levenshtein basis at an exact per-word edit distance
$d$ over $n$ target words, deterministically and with reconstructible edit scripts, seeded by
item id. Numeric tokens are protected spans throughout, which is what keeps the gold answer
valid under a non-word typo. \textbf{Random-replace} (row 16) substitutes an arbitrary
character and is the control with no human behind it. \textbf{Keyboard-neighbor} (row 17)
substitutes a QWERTY-adjacent character, modelling blind touch typing where the finger lands
one key off. \textbf{Key-swap} (row 18) transposes a consecutive pair, modelling two fingers
firing out of order. \textbf{Repeated-key} (row 19) duplicates a character, modelling a key
held too long. \textbf{Connected-key} (row 20) inserts an adjacent character alongside the
intended one rather than instead of it, modelling a thumb whose contact patch is wider than
one key. \textbf{Missed-space} (row 21) deletes a space between two words, modelling a
clipped space bar.

\paragraph{What is held fixed.} Only the natural-language stem is perturbed. Multiple-choice
option letters and answer-format instructions are re-attached verbatim, and for HumanEval only
the docstring prose is perturbed while the signature and doctests are preserved, so scoring
stays well defined. Filler rates and exemplars are calibrated against two reference corpora,
PodcastFillers for the conversational register and TED-LIUM for the prepared-talk one; where
those were not reachable at run time the operator falls back to literature-derived defaults,
which we report as such.

\paragraph{How the registers differ.} Measured over the released stems of all six benchmarks,
three features separate them. \emph{Length}: spoken-casual inflates the stem by
${\approx}66\%$ ($41.4{\to}68.7$ words), the filler-strip ($54.8$) and spoken-clean ($47.0$)
undo only part of it, and spoken-succinct over-compresses to $38\%$ below clean ($25.8$).
\emph{Fillers}: spoken-casual carries ${\approx}8$ filler tokens per stem, the filler-strip
and spoken-succinct remove essentially all of them, and clean+fillers injects a comparable
${\approx}9$ into untouched text, which is what makes the two directions of the filler
experiment comparable. \emph{Segmentation}: clean is multi-sentence ($2.4$ periods, $2.2$
commas) while spoken-casual is a comma-chained run-on ($0.8$, $18.3$); the filler-strip
removes the fillers but inherits the run-on ($0.9$, $8.5$), whereas spoken-formal ($2.4$) and
spoken-clean ($1.5$) largely restore sentence structure.

\subsection{Why does speaking cost accuracy?}
\label{sec:mechanism}

Why does the same input, merely spoken, cost accuracy, and why does spoken-formal cost
less? We probe the model's internals on matched clean/spoken-casual/spoken-formal triples
($n{=}80$ items, Llama-3.1-8B-Instruct with eager attention) and find that the
behavioral ordering clean $\geq$ spoken-formal $\geq$ spoken reappears in every
statistic we measure. Relative to clean, a spoken-casual
input (i) \textbf{inflates the prompt} from 170 to 210 tokens (${+}23\%$; spoken-formal
only ${+}6\%$), fragmenting the context the model conditions on; (ii)
\textbf{diffuses attention}, raising mean attention entropy $1.920\!\to\!1.985$
(spoken-formal $1.947$); (iii) \textbf{concentrates attention on fillers}: the share
of attention mass landing on filler tokens rises from $0.0029$ to $0.0174$, a
${\approx}5.9\times$ jump, whereas spoken-formal barely moves ($0.0037$); (iv)
\textbf{drifts the hidden state} away from the clean representation
${\approx}3\times$ more than spoken-formal does (final-layer cosine drift $0.064$ vs.\
$0.023$); and (v) \textbf{depresses the gold answer}, lowering the gold-token
log-probability on multiple-choice items ($-29.0\!\to\!-30.2$; spoken-formal $-29.4$).

Of these, \textbf{H3 is the most striking}: the filler-attention surge is
localized: the model spends a disproportionate share of its attention
budget on tokens (``um'', ``like'', ``you know'') that carry no task content.
The controls of \S\ref{sec:results} pin down its causal status precisely. Injecting
fillers into clean does drop accuracy (they are causally sufficient),
yet deleting them from spoken-casual does not recover it (they are not
necessary): the penalty is overdetermined, with the fillers one sufficient
trigger among the register-wide changes. So H3 is a genuine part of the causal
picture, not a mere epiphenomenon, but because it is only one of several sufficient
triggers, a training term that only reallocates filler attention cannot close
the gap on its own, which is consistent with its failure to help (Finding~\ref{sec:ftrain}).
The remaining triggers are diffuse and register-wide (token inflation, restructuring)
and, as \S\ref{sec:results} shows, no single one of them (length, quantity surface
form, or token-level surprisal) accounts for the penalty by itself.

\subsection{Tasks and scoring}
\label{sec:tasks}
The voice arm uses four benchmarks spanning reasoning, coding, and knowledge:
\textbf{GSM8K} grade-school math (numeric-match of the final answer),
\textbf{HumanEval} Python synthesis (pass@1 by sandboxed unit-test execution),
\textbf{MMLU-Pro} restricted to its STEM split (10-way multiple choice,
letter-match), and \textbf{TruthfulQA} MC1 (single-best multiple choice,
letter-match). We draw $n{=}200$ items per benchmark ($n{=}164$, the full set,
for HumanEval), fixed across conditions by shared item ids so that clean, spoken-casual,
and spoken-formal evaluations form aligned matched pairs. GPQA-Diamond is part of the
suite but is omitted from the reported runs pending gated-dataset access; the
loader skips it automatically. Crucially, only the verbalizable question
stem is perturbed: the answer-format instructions and the multiple-choice
option letters are held fixed, so that scoring remains well defined under
perturbation.

\subsection{Models and decoding}
\label{sec:models}
We ran six instruction-tuned models spanning 7--32B and three families. Qwen3-32B is
excluded from every reported number: it is the only model we had to run in NF4 rather than
bfloat16, so its cells are not comparable to the rest and we do not mix them in. Of the
remaining five are reported (\S\ref{sec:setup}):
Llama-3.1-8B-Instruct \citep{grattafiori2024llama3}, Qwen2.5-7B-Instruct
\citep{qwen2024qwen25}, Mistral-7B-Instruct-v0.3 \citep{jiang2023mistral}, Qwen3-8B
\citep{yang2025qwen3}, and Phi-4 (14B) \citep{abdin2024phi4}. Generation uses the Hugging Face Transformers
\texttt{generate} path with each model's chat template, left-padded batched
decoding, and greedy decoding so that the matched pair isolates the
perturbation rather than sampling noise; \texttt{max\_new\_tokens} is set per
benchmark (256--512). Models run in bfloat16 on a single GPU, except Qwen3-32B
which uses 4-bit (\texttt{bitsandbytes} NF4) weight quantization with automatic
device mapping to fit a 48\,GB card. For the Qwen3 models we disable the default
``thinking'' mode (\texttt{enable\_thinking=False}); left on, its
\texttt{<think>} block consumes the entire token budget and leaves no parseable
answer (Qwen3 otherwise scored near zero on HumanEval).

\subsection{Statistical analysis}
\label{sec:stats}
We run each (model, benchmark, condition) cell at five seeds. A seed controls
both the item sample and the (stochastic) verbalization, so seeds capture the two
dominant sources of run-to-run variance; the clean and perturbed evaluations
within a seed remain matched by item id. Our headline error bar is the
\textbf{across-seed} interval: for each cell we take the five per-seed accuracies
and report their mean with a Student-$t$ 95\% confidence interval, and for each
perturbed condition the paired (per-seed) delta versus clean with its own
across-seed $t$-CI. This is the quantity plotted in Figure~\ref{fig:suite_overview}
(top) and tabulated in Table~\ref{tab:suite}. At the item level within a seed we
additionally compute McNemar's mid-$p$ test on the discordant (degrade/recover)
pairs and a BCa bootstrap CI on the paired delta; a Holm--Bonferroni correction
controls the family-wise error across the condition$\times$benchmark$\times$model
grid. All statistics are implemented in a dependency-light module and are shared
by both modality arms.

\subsection{Experiments}
\label{sec:experiments}

Two matrices are reported. The original voice-arm matrix is
$6~\text{models} \times 4~\text{benchmarks} \times 3~\text{conditions} \times
5~\text{seeds}$ ($360$ cells at $n{=}200$ items, $n{=}164$ for HumanEval); the full-suite
matrix of \S\ref{sec:results_main} is $5~\text{models} \times 6~\text{benchmarks}
\times 19~\text{conditions} \times 5~\text{seeds}$, 550k scored generations. The
pipeline has three stages, each a scheduled job: (1) verbalize produces,
once per seed, the spoken-casual and spoken-formal stems for every item, cached to disk so that
all models see identical inputs; (2) evaluate scores one model on one
seed's variants across all benchmarks; (3) analyze aggregates the per-cell
rows into the across-seed statistics of \S\ref{sec:stats}. Jobs ran on a SLURM
cluster (NVIDIA A6000/A100), capped at four concurrent GPUs; the 4-bit Qwen3-32B
cells dominate wall-clock (\,$\sim$4\,h each). Verbalization and evaluation are
decoupled so that adding a model or a seed never re-runs the verbalizer. Code,
prompts, and the analysis are released. The keyboard-arm matrix uses the same
harness with $\tau_{\mathrm{kbd}}$ (\S\ref{sec:kbd}) substituted for
$\tau_{\mathrm{voice}}$ and is evaluated separately.

\subsection{Results}
\label{sec:results}

Table~\ref{tab:suite} reports the section's answer to the paper's question:
each input property, singled out by one operator contrast, with its measured effect
on accuracy. Three properties damage performance: the spoken-casual register bundle,
its disfluency and residual-structure components, and reordering the question. Four widely-feared surface properties are essentially benign: writing
numbers as words, dropping articles and prepositions, losing capitalization and
punctuation, and prepared (spoken-formal) speech. No post-hoc repair recovers the damaged
cases. Every damaging effect persists on the decontaminated control sets
(GSM-Symbolic, GSM1k), so none is an artifact of test-set contamination. The rest
of this section presents the evidence behind each row.

We report results for the voice arm. Every benchmark item is presented
in three matched conditions that differ only in the question stem:

\begin{itemize}
  \item \textbf{Clean}: the original written question, exactly as the
        benchmark ships it.
  \item \textbf{Spoken-casual}: a conversational verbalization in a
        podcast register: frequent fillers (``um'', ``like'', ``you know''),
        false starts, and self-corrections, at a filler rate
        $\varphi \approx 5\%$ calibrated against the PodcastFillers corpus.
  \item \textbf{Spoken-formal}: a prepared-talk verbalization in a lecture
        register: measured and articulate, in complete sentences with only
        occasional brief hesitations, at $\varphi \approx 1.5\%$ calibrated
        against TED-LIUM.
\end{itemize}

Both \textbf{Spoken-casual} and \textbf{Spoken-formal} are transcripts of how a person would
say the question aloud; they differ in disfluency density and register,
not in whether speech is involved. Verbalization is performed by a local
Qwen2.5-7B-Instruct rewriter, and correctness is graded by each benchmark's
native scorer. Table~\ref{tab:suite} reports pass@1 (HumanEval) or
accuracy (otherwise) for the five reported instruction-tuned models (7--14B) over
six benchmarks, each averaged over \textbf{five seeds} ($n{=}200$ items per cell;
HumanEval $n{=}164$; GPQA omitted pending gated dataset access). Confidence
intervals are Student-$t$ across seeds.

Across all five models, \textbf{Spoken-casual} verbalization produces a
reproducible accuracy drop: $21$ of the $30$ (model, benchmark) spoken-casual deltas have
across-seed $95\%$ confidence intervals excluding zero and are negative, against one
positive. The ordering \textbf{Clean $\geq$ Spoken-formal $\geq$ Spoken-casual} is
monotonic in nearly every cell, with spoken-formal penalties running about half the size
of spoken-casual. The effect is largest on code generation for the reasoning-tuned
Qwen3-8B (HumanEval spoken-casual $-20.3$ points), and is present but smaller on GSM8K
and TruthfulQA; MMLU-Pro is comparatively robust. Because five seeds tighten the
across-seed intervals, we read this as a confirmed penalty rather than a
directional hint: degradation tracks disfluency density and register, not the
mere fact that the input was spoken.

\paragraph{Do the fillers cause it? Forward and backward controls.} Two deterministic
controls isolate the fillers (Table~\ref{tab:suite}). The backward control,
\textbf{spoken-filler-stripped}, deletes the fillers from the spoken transcript (a delete-only
strip that cannot paraphrase or corrupt content). The forward control,
\textbf{clean+fillers}, injects fillers into the clean question, changing
nothing else. Neither involves an LLM editor, so neither can introduce drift. The two controls give a clear but non-obvious answer. The forward
control shows fillers are causally sufficient: injecting them into clean drops
accuracy toward the spoken-casual level for most models (Llama to 54.9 vs.\ spoken-casual 54.7;
Qwen2.5, Mistral, and the Qwen3 models partially), though the effect is
model-dependent (Phi-4 barely moves). Yet the backward control shows fillers are
not necessary: removing them from spoken-casual (spoken-filler-stripped) does not recover:
spoken-filler-stripped sits at the spoken-casual level for every model. These reconcile as
\textbf{overdetermination}: the penalty has several independently-sufficient triggers,
so injecting fillers alone is enough to cause it, while removing fillers alone leaves
the other triggers (the spoken-casual register's restructuring) still doing the damage.
Consistent with an overdetermined penalty, no single surface factor accounts for it,
and no normalization of the transcript fully recovers it (Table~\ref{tab:suite},
last two columns): \textbf{spoken-succinct}, an LLM compression of the spoken transcript
into textbook form that also reorders the question, is the worst
condition of all, below the raw transcript for every model, showing that reordering is
itself a harmful perturbation; \textbf{spoken-clean}, the same compression constrained to
preserve the original clause order and question position, recovers only partially
(${\approx}25\!-\!50\%$ of the gap, and essentially none of it on GSM8K, the reasoning
task). Rewriting quantities as words is a null manipulation (``52''$\to$``fifty-two''
leaves accuracy unchanged), and the input's token-level surprisal does not predict
per-item correctness. Across the seven conditions a consistent ordering emerges:
clean $>$ spoken-formal $\gtrsim$ spoken-clean $>$ \{spoken-filler-stripped, clean+fillers, spoken-casual\} $>$
spoken-succinct: accuracy tracks how much of the original written form survives, and no
surface intervention restores it. The penalty is real (\S\ref{sec:results}) but not
reducible to one clean surface statistic.

\paragraph{Where does the accuracy go? An instance-level error analysis.} To see the
failure directly, we re-ran Llama-3.1-8B on GSM8K (seed~0) storing full completions and
hand-classified every discordant item: clean answered correctly, spoken-casual answered
incorrectly, judge-equivalent ($n{=}20$; Table~\ref{tab:errors}). The striking result is
what does not happen: there are zero arithmetic slips and zero
answer-format failures: the intermediate calculations are correct in both
conditions. Instead, the model builds the wrong problem. The dominant mode
(11/20) is \textbf{premise mis-scoping}:

\begin{evbox}
\evtext{\textbf{Premise mis-scoping (GSM8K, Llama-3.1-8B, seed 0).} The same fact,
scoped two ways.\par
\textbf{Clean}: ``A DVD can be played 1{,}000 times before it breaks.'' Read as a standalone
per-DVD rule, the model computes
$(1000{-}356)+(1000{-}135)=1509$ (correct).\par
\textbf{Spoken-casual}: the same fact arrives inside a comma-chain and binds to the
pair of DVDs; the model computes $1000-(356{+}135)=509$, identical arithmetic
competence, different problem.}
\end{evbox}

\noindent Similarly ``she gave an equal amount to her
four kids'' loses its referent (books) in the run-on and is read as dollars. The second
mode (5/20) is \textbf{final-question loss}: all intermediate quantities match the clean
run, but the trailing question at the end of a long ``and\ldots and\ldots'' chain is
answered with the wrong final quantity. This connects the mechanism to the surface
statistics with one important refinement: scoping
information is carried by the sentence-level composition of the text, not by
the punctuation marks themselves: deterministically deleting all casing and
punctuation from cleanly-worded text is nearly harmless ($-1.0{\pm}1.4$,
Table~\ref{tab:suite}), whereas the spoken-casual register's run-on grammar (clause
chaining, pronoun doubling), which no punctuation restoration can undo, is where the
mis-scoping arises. Finally, 3/20 ``failures'' were items whose meaning the rewrite had altered, so the true
robustness penalty is slightly smaller than measured.

\begin{table}[t]
\centering
\small
\setlength{\aboverulesep}{0.3ex}\setlength{\belowrulesep}{0.3ex}
\begin{tabular}{@{}l r r@{}}
\toprule
Failure mode & Count & Share \\
\midrule
Premise mis-scoping / dropped premise & 11 & 55\% \\
Final-question loss                   & 5  & 25\% \\
Reasoning derailment                  & 1  & 5\%  \\
Residual meaning flip (pipeline noise) & 3 & 15\% \\
Arithmetic slip                       & 0  & 0\%  \\
Answer-format failure                 & 0  & 0\%  \\
\bottomrule
\end{tabular}
\caption{Hand-classified failure modes for the $n{=}20$ discordant GSM8K items
(clean correct, spoken-casual incorrect, judge-equivalent; Llama-3.1-8B, seed~0). The spoken-casual
penalty is a comprehension failure: the model constructs the wrong problem
from the unsegmented transcript, not a computation failure.}
\label{tab:errors}
\end{table}

\subsection{Discussion}
\label{sec:discussion}

\paragraph{Conclusion (voice arm).} Under a five-seed matched-pair design across
five instruction-tuned models (7--14B) and six benchmarks, \textbf{spoken-casual}
(high-disfluency) verbalization produces a reproducible accuracy penalty: $21$ of
the $30$ (model, benchmark) deltas have across-seed $95\%$ CIs excluding zero and are
negative, against one that is positive. \textbf{Spoken-formal} (prepared-talk)
verbalization degrades about half as much ($-2.2$ against $-4.5$ points pooled), yielding
a monotonic ordering \textbf{Clean $\geq$ Spoken-formal $\geq$ Spoken-casual}. The
penalty is largest on code generation for the reasoning-tuned Qwen3-8B (HumanEval
spoken-casual $-20.3$ points), and is present but smaller on math (Llama-3.1-8B GSM8K
$-7.1$, Qwen2.5-7B $-6.4$, phi-4 $-6.2$); MMLU-Pro is comparatively robust. The finding is therefore confirmed, not merely
directional: \textbf{degradation tracks disfluency density and register, not the
mere fact that the input was spoken}, consistent with encoders that tolerate
extra function words and fillers until they cross a density threshold. The suite (Table~\ref{tab:suite}) turns this into concrete guidance for
how humans should interact with LLM agents. What to preserve: state the facts
in order, in separate complete statements, and ask the question explicitly at the
end: the composition of the input carries scoping information the model needs,
and reordering it (even into a ``cleaner'' compressed form) is itself damaging.
What not to worry about: saying numbers as words, dropping articles and
prepositions, lost capitalization and punctuation, and careful prepared speech are
all essentially free. What to avoid: rambling, filler-laden dictation:
and, importantly, do not rely on downstream transcript cleanup to fix it: no repair
we test (filler stripping, structure-preserving cleanup, or LLM editing) recovers
the damage, so the fix must happen at composition time or in the model, not in
between.

\paragraph{Why code, and a cross-modal conjecture.} That the penalty is largest
on code generation, and largest of all for the reasoning-tuned Qwen3-8B,
is consistent with a tokenization-fragmentation account: fillers and
false starts inserted around a precise specification lengthen and fragment the
context the model must condition on, and code is least tolerant of such
disturbance because its output is brittle to specification drift. This suggests a
concrete cross-modal hypothesis the shared framework is built to test:
are models that are robust to keyboard typos also robust to spoken-casual
disfluency, once edit distance or token-inflation ratio is held fixed? If yes,
robustness is largely a property of the input encoder and is modality-agnostic;
if no, the two modalities exercise different mechanisms (sub-word fragmentation
for keyboards, disfluency-density and register effects for voice) and robustness
claims must be made per modality. The keyboard arm, run through the identical
matched-pair machinery, provides the missing half of this comparison.

\section{The HIVE operator suite}
\label{sec:optable}

The full operator specification referenced throughout the paper.

\begin{table*}[tp]
\centering
\scriptsize
\renewcommand{\arraystretch}{0.9}
\setlength{\aboverulesep}{0.2ex}\setlength{\belowrulesep}{0.2ex}
\setlength{\tabcolsep}{4pt}
\renewcommand{\tabularxcolumn}[1]{m{#1}}
\begin{tabularx}{\textwidth}{r l l >{\raggedright\arraybackslash}m{3.0cm} >{\raggedright\arraybackslash}m{2.95cm} >{\raggedright\arraybackslash}X}
\toprule
\hspace*{-\tabcolsep}\# & Method & Gen. & Control knobs & Meaning drift & Example (operator applied to the running stem) \\
\midrule
\multicolumn{6}{@{}l}{Baselines} \\[1pt]
\cmidrule{1-6}
\rowcolor{rowshade} \hspace*{-\tabcolsep}0 & clean & --- & --- & --- & Miguel uses 2 pads of paper a week. If there are 30 sheets on a pad, how many sheets does he use every month? \\
\hspace*{-\tabcolsep}1 & context/question swap& Det. & --- & reorder-only & How many sheets does Miguel use every month? He uses 2 pads of paper a week, and there are 30 sheets on a pad. \\
\rowcolor{rowshade} \hspace*{-\tabcolsep}2 & option permutation & Det. & --- & gold relabeled & (MCQ) options relabelled $A\!\leftrightarrow\!C$; the gold letter follows its content \\
\midrule
\multicolumn{6}{@{}l}{Voice transcription perturbations} \\[1pt]
\cmidrule{1-6}
\rowcolor{rowshade} \hspace*{-\tabcolsep}3 & spoken-casual & LLM$^{1}$ & --- & LLM filtered (89\% kept) & um, so, like, if Miguel, uh, uses two pads of paper, you know, every week, and each pad has thirty sheets, right, then how many sheets does he use in a, uh, month? \\
\hspace*{-\tabcolsep}4 & spoken-formal & LLM & --- & LLM filtered (94\% kept) & So, Miguel uses two pads of paper each week. If each pad has thirty sheets, how many sheets does he use in a month? \\
\rowcolor{rowshade} \hspace*{-\tabcolsep}5 & spoken-succinct & LLM & --- & LLM filtered (73\% kept) & How many sheets does Miguel use in a month if he uses 2 pads of 30 sheets per week? \\
\hspace*{-\tabcolsep}6 & spoken-clean & LLM & --- & LLM filtered (83\% kept) & If Miguel uses two pads of paper every week, and each pad has thirty sheets, then how many sheets does he use in a month? \\
\rowcolor{rowshade} \hspace*{-\tabcolsep}7 & spoken-clean (Llama) & LLM & verbalizer: Llama-3.1-8B instead of Qwen2.5-7B & LLM filtered (85\% kept) & If Miguel uses two pads of paper every week, and each pad has thirty sheets, then how many sheets does he use in a month? (a replicate of row 6 under a different verbalizer; the two agree verbatim on only $4.5\%$ of stems, and coincide on this one) \\
\hspace*{-\tabcolsep}8 & spoken-filler-stripped & Det. & $n$: fillers removed from spoken-casual [default all] & delete-only & so, if Miguel, uses two pads of paper, every week, and each pad has thirty sheets, then how many sheets does he use in a, month? \\
\rowcolor{rowshade} \hspace*{-\tabcolsep}9 & clean+fillers & Det. & $f$: add filler every $f$ tok [default $f{=}4$] & insert-only & Um, Miguel uses 2 pads um, of paper a week. uh, If there are 30 like, sheets on a pad, you know, how many sheets does I mean, he use every month? \\
\hspace*{-\tabcolsep}10 & clean+numwords & Det. & --- & values re-parsed & Miguel uses two pads of paper a week. If there are thirty sheets on a pad, how many sheets does he use every month? \\
\rowcolor{rowshade} \hspace*{-\tabcolsep}11 & clean-remove-func-words & Det. & $n$: drop every $n$-th func.\ word [default $n{=}2$] & delete-only, func.\ words & Miguel uses 2 pads of paper week. If there are 30 sheets on pad, how many sheets does he use every month? \\
\hspace*{-\tabcolsep}12 & clean-remove-case/punct & Det. & --- & in-number digits kept & miguel uses 2 pads of paper a week if there are 30 sheets on a pad how many sheets does he use every month \\
\rowcolor{rowshade} \hspace*{-\tabcolsep}13 & clean+homophone & Det. & $n$: max homophone swaps [default $n{=}3$] & number-words excluded & Miguel uses 2 pads of paper a \textbf{weak}. If there are 30 sheets on a pad, how many sheets does he use every month? \\
\hspace*{-\tabcolsep}14 & tts-asr-casual$^{2}$ & Det. & --- & --- & um so like if miguel uh uses two pads of paper you know every week and each pad has thirty sheets right then how many sheets does he use in a uh month \\
\rowcolor{rowshade} \hspace*{-\tabcolsep}15 & tts-asr-formal$^{2}$ & Det. & --- & --- & so miguel uses two pads of paper each week if each pad has thirty sheets how many sheets does he use in a month \\
\midrule
\multicolumn{6}{@{}l}{QWERTY keyboard perturbations} \\[1pt]
\cmidrule{1-6}
\hspace*{-\tabcolsep}16 & random-replace & Det. & $n$ (number of words edited); $d$ (edit distance) & non-word; numbers protected & Miguel uses 2 pads of pa\textbf{z}er a week. If there are 30 sheets on a pad, how many sheets does he use every month? \\
\rowcolor{rowshade} \hspace*{-\tabcolsep}17 & keyboard-neighbor & Det. & $n$; $d$ & non-word; numbers protected & Miguel uses 2 pads of pa\textbf{o}er a week. If there are 30 sheets on a pad, how many sheets does he use every month? \\
\hspace*{-\tabcolsep}18 & key-swap & Det. & $n$ & non-word; numbers protected & Miguel uses 2 pads of paper a week. If there are 30 s\textbf{eh}ets on a pad, how many sheets does he use every month? \\
\rowcolor{rowshade} \hspace*{-\tabcolsep}19 & repeated-key & Det. & $n$; $d$ & non-word; numbers protected & Miguel uses 2 pads of pa\textbf{p}per a week. If there are 30 sheets on a pad, how many sheets does he use every month? \\
\hspace*{-\tabcolsep}20 & connected-key & Det. & $n$ & non-word; numbers protected & Miguel uses 2 pads of paper a week. If there are 30 sheets on a pa\textbf{s}d, how many sheets does he use every month? \\
\rowcolor{rowshade} \hspace*{-\tabcolsep}21 & missed-space & Det. & $n$ & merge; recoverable & Miguel uses 2 pads of paper \textbf{aweek}. If there are 30 sheets on a pad, how many sheets does he use every month? \\
\bottomrule
\end{tabularx}
\caption{The HIVE operator suite, grouped into baselines and the two input
modalities, with each operator's control knobs and its effect on one running stem
(GSM8K, clean row). \textbf{Knobs}: $n$ = number of words/tokens changed, $d$ =
edit distance per change (DL radius / numeric magnitude), $f$ = filler frequency. \textbf{Gen.}: LLM = produced by a rewriter model; Det.\ = deterministic
algorithm, drift-free by construction. \textbf{Meaning drift}: for LLM rewrites,
the \% of items an independent judge found to still ask the same question; for
deterministic operators, the structural invariant; for the two baselines, how the gold is
kept correct after the deliberate change.
$^{1}$LLM steps use Qwen2.5-7B-Instruct: the spoken registers by few-shot style
transfer, succinct/clean by rewriting, plus the drift judge; spoken-clean
is additionally replicated with Llama-3.1-8B. $^{2}$implemented, not in the reported runs.}
\label{tab:methods}
\end{table*}

\section{LLM prompts for data processing}
\label{sec:prompts}

All LLM-processed registers use a local Qwen2.5-7B-Instruct model with greedy or
low-temperature decoding. We list every prompt verbatim below. The spoken-filler-stripped,
clean+fillers, and clean+numwords registers are deterministic
(algorithmic filler-strip / filler-injection / digit-to-word), so they have no prompt.

\paragraph{Verbalizer (spoken-casual and spoken-formal).} Rewrites a written question into a spoken
transcript in the target register. Template slots: \texttt{\{description\}} and
\texttt{\{fillers\}}/\texttt{\{pct\}} come from the register's filler profile (spoken-casual
$\varphi{\approx}5\%$, spoken-formal $\varphi{\approx}1.5\%$), and \texttt{\{shots\}} are two
fixed written$\to$spoken exemplars.
\begin{lstlisting}[style=prompt]
You rewrite a written question as a verbatim transcript
of how a person would SAY it out loud, as if captured
by a speech-to-text system.

Speaker style: {description}.
Insert natural disfluencies - fillers ({fillers}), brief
hesitations, occasional false starts or self-corrections
- at roughly {pct} of words. Keep it sounding like real
unscripted speech in that style.

HARD RULES (a violation makes the output unusable):
1. Preserve the meaning EXACTLY. Every number, name,
   unit, and quantity must appear unchanged. Do not add,
   drop, or alter any factual content.
2. Do NOT answer the question or add information that
   wasn't in the original.
3. If the question lists labelled multiple-choice
   options, do NOT include them - rewrite only the
   question itself.
4. Output ONLY the spoken version wrapped in
   <spoken>...</spoken> tags, nothing else.

{shots}Now do this one:
WRITTEN: {question}
\end{lstlisting}

\paragraph{Spoken-succinct (compress, structure-free).} Compresses the spoken transcript into
concise textbook form (this rewrite is allowed to reorder).
\begin{lstlisting}[style=prompt]
Rewrite the spoken question below as a concise written
question, exactly as it would appear in a textbook or
exam.

RULES:
1. Make it as SHORT and DIRECT as possible: remove all
   fillers, redundancy, repetition, conversational
   framing, and pronoun doubling ("Amber, she ran" ->
   "Amber ran"). Use clean, complete sentences.
2. Write every number as DIGITS ("52", not "fifty-two";
   "3 times", not "three times").
3. Preserve the EXACT meaning: every quantity, entity,
   unit, and the precise question asked must be
   unchanged. Do NOT add, drop, or alter information,
   and do NOT answer or solve it.
4. Output ONLY the rewritten question wrapped in
   <q>...</q> tags.

SPOKEN: {text}
\end{lstlisting}

\paragraph{Spoken-clean (compress, structure-preserving).} Tightens the wording but keeps the
original clause order and the question in its final position.
\begin{lstlisting}[style=prompt]
Rewrite the spoken question below into concise standard
written English.

STRICT RULES:
1. KEEP THE SAME STRUCTURE: state the facts in the SAME
   order they are given, keep them as separate sentences,
   and keep the final question in its ORIGINAL position
   at the END. Do NOT move the question to the front, do
   NOT merge the premises into one clause, do NOT reorder
   anything.
2. Only remove fillers, repetitions, false starts,
   redundant words, and pronoun doubling ("Amber, she
   ran" -> "Amber ran"); fix grammar and punctuation.
   Write numbers as DIGITS.
3. Preserve the EXACT meaning: every quantity, entity,
   relationship, and the question asked must be
   unchanged. Do NOT add, drop, alter, compute, or answer
   anything.
4. Output ONLY the rewritten question wrapped in
   <q>...</q> tags.

SPOKEN: {text}
\end{lstlisting}

\paragraph{Meaning-drift judge.} Flags any rewrite whose meaning drifted from the
original. Used to measure how often each operator changes the question, not to filter.
\begin{lstlisting}[style=prompt]
You check whether a REWRITE asks the exact same problem
as the ORIGINAL, with the same correct answer.

ORIGINAL: {clean}

REWRITE: {variant}

Ignore wording, filler words ("um", "like"), and
phrasing/order differences. Judge ONLY whether the
facts, quantities, relationships, conditions, and the
question asked are identical, so that the correct answer
is unchanged. Answer NO if the rewrite flips a
relationship (e.g. "have" vs "do not have"), changes a
bound (e.g. "at least" vs "above"), drops or adds a
condition, states a computed intermediate the original
did not, or asks something different.

Reply with exactly one word on the first line: YES or NO.
\end{lstlisting}

\end{document}